# Facilitating reflection in teletandem through automatically generated conversation metrics and playback video

Aparajita Dey-Plissonneau[1], Hyowon Lee[2], Michael Scriney[3], Alan F. Smeaton[4], Vincent Pradier[5], Hamza Riaz[6]

**Abstract**

This pilot study focuses on a tool called L2L that allows second language (L2) learners to visualise and analyse their Zoom interactions with native speakers. L2L uses the Zoom transcript to automatically generate conversation metrics and its playback feature with timestamps allows students to replay any chosen portion of the conversation for post-session reflection and self-review. This exploratory study investigates a seven-week teletandem project, where undergraduate students from an Irish university learning French (B2) interacted with their peers from a French university learning English (B2+) via Zoom. The data collected from a survey (N=43) and semi-structured interviews (N=35) show that the quantitative conversation metrics and qualitative review of the synchronous content helped raise students' confidence levels while engaging with native speakers. Furthermore, it allowed them to set tangible goals to improve their participation, and be more aware of what, why, and how they are learning.

## 1. Introduction

In the context of the Covid-19 pandemic, the possibilities of enhancing the learning and teaching experience by developing the affordances of online learning platforms has been a pressing need. In foreign language education, videoconferencing platforms afford connecting with native speakers in geographically distant institutions for language and intercultural competences development via telecollaboration (Dooly & O'Dowd, 2018). Facilitated by technological progress, innovative learning activities, such as students reviewing the recordings of their online interactions for reflection and learning, are gaining importance in language pedagogy (Rivière & Guichon 2014). This pilot study describes a tool that we conceived and developed in Dublin City University (DCU) called the L2L platform (Dey-Plissonneau, Lee, Pradier, Scriney, Smeaton, 2021) that automatically generates conversation metrics and content visualisation of L2 learners' synchronous Zoom teletandem interactions with native speakers.

As part of this tandem project that involved seven weekly sessions in the second semester of the 2020-2021 academic year, 60 students in DCU learning French (B2, 20-21 years, mixed gender) interacted via Zoom with 45 students in Paris Sciences et Lettres (PSL) learning English (B2+, 19-20 years, mixed gender). DCU students majored in business, media, law and translation studies, while PSL students majored in humanities and sciences. The objectives of these interactions were to develop their second language (L2) knowledge, interaction skills, and intercultural competences. Various challenges encountered in previous teletandem projects led us to the creation of the L2L platform. Firstly, high attrition rates in the context of online learning, especially as monitoring synchronous interactions on a weekly basis is a tall order for lecturers. Secondly, the fast-pace of synchronous interactions in the target language involves high cognitive load, anxiety, and code-switching. These need to be complemented by asynchronous reflection as the synchronous interactions are too ephemeral for retention. Finally, intermediate students often feel that they have reached a plateau in their learning progress (Richards, 2008), and thus need to set very specific learning objectives.

---

[1] Dublin City University, Dublin, Ireland; *aparajita.dey-plissonneau@dcu.ie*; https://orcid.org/0000-0003-1429-6861

[2] Dublin City University, Dublin, Ireland; *hyowon.lee@dcu.ie*; https://orcid.org/0000−0003−4395−7702

[3] Dublin City University, Dublin, Ireland; *michael.scriney@dcu.ie*; https://orcid.org/0000−0001−6813−2630

[4] Dublin City University, Dublin, Ireland; *alan.smeaton@dcu.ie*; https://orcid.org/0000−0003−1028−8389

[5] Paris Sciences et Lettres Université, Paris, France; *vincent.pradier@psl.eu*; https://orcid.org/0000-0002-7050-6408

[6] Dublin City University, Dublin, Ireland *hamza.riaz2@mail.dcu.ie*; https://orcid.org/0000-0001-6339-6194

The L2L platform uses Zoom's automated transcript facility to create conversation metrics for each interaction session, at the individual and inter-individual levels. Figure 1 shows the L2L dashboard right after a conversation was held between two DCU and two PSL students. Students can visualise their conversation share and the conversation flow between all the participants of their group. Additionally, a video playback timeline with colour-coded audio graphs allows one to select any point on the timeline and play the recording from that point onward. These affordances allow students to review (self- and peer-review) their otherwise ephemeral synchronous exchanges both at the quantitative and qualitative levels.

Figure 1: Dashboard of the L2L platform

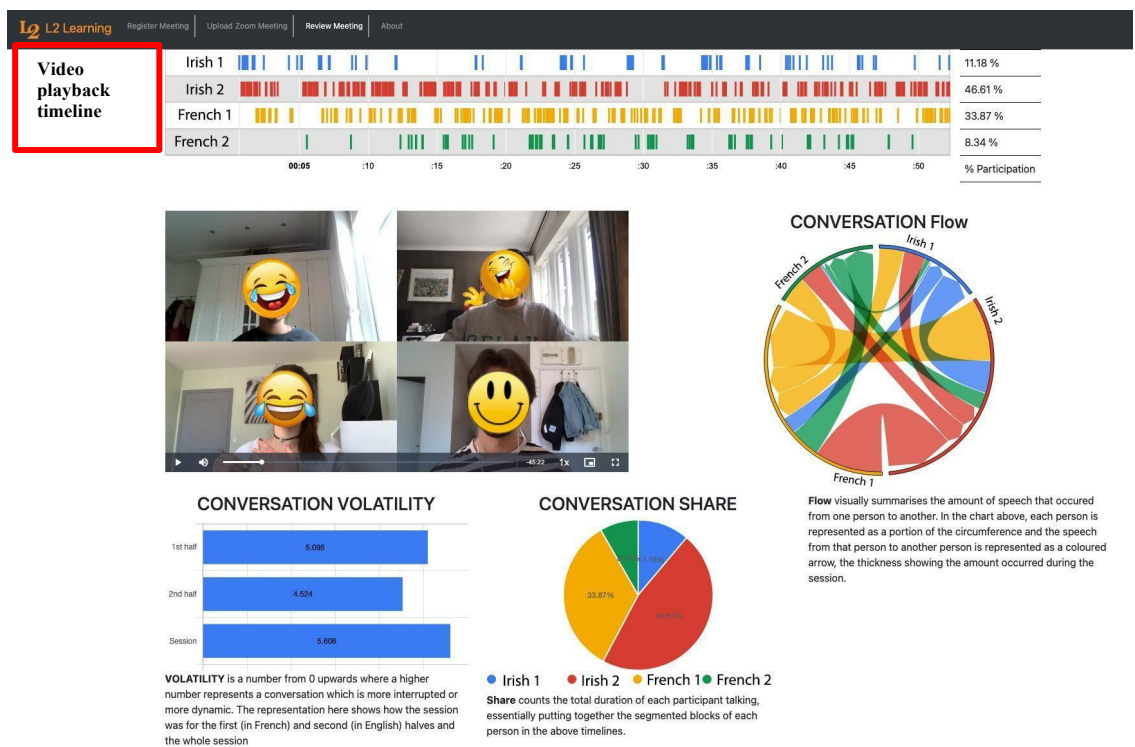

This study examines the principles for the pedagogical design that are generated, as successful learners are those who are "aware of their perceptions, attitudes, and abilities and are knowledgeable about the learning process" (Hauck, p. 2005). Introducing the L2L platform tool in the pedagogical design of the teletandem project led us to focus on the following research questions.

- What is the added value of using participation visualisations on students' learning via teletandem?
- What principles for the pedagogical design are generated?

## 2. Method

The unit of analysis in this exploratory study is the participating DCU students who were engaged in self-review/reflection using the L2L platform and had participated in two other teletandem projects in the previous semesters without L2L. The data used includes the results of a survey taken by 43 DCU students and semi-structured interviews of 35 DCU students gathered after the end of the seven-week teletandem project using L2L. Our thematic analysis of the interviews sheds light on the correlation between the learners' perception of the L2L visualisations and their influence on the L2 learning process.

The pedagogical design for the teletandem comprised weekly activities conducted in three phases (Mayer, 2005):
- asynchronous pre-session written interactions on forums;
- synchronous conversation-based videoconferencing in triads (2 DCU and 1 PSL) and quadruplets (2 DCU and 2 PSL);
- post-session individual reflections on their interactions using L2L.

The asynchronous forum and synchronous Zoom interactions were based on topics related to personal interests and university life in the first two sessions, and then moved on to current affairs such as statelessness, racism, state violence, geo-politics, and the treatment of these debates in French and Irish/English newspapers, in line with their respective module curricula. Four post-session quantitative and qualitative reflections using L2L were conducted by DCU students with the aim to review their strengths and weaknesses in effectively interacting with native speakers.

## 3. Results and discussion

In response to the survey question whether graphs and video playback were helpful to analyse their post-session reflection and learning, 76.7% of respondents replied 'yes' (see Figure 2).

Figure 2: Learner feedback on L2L

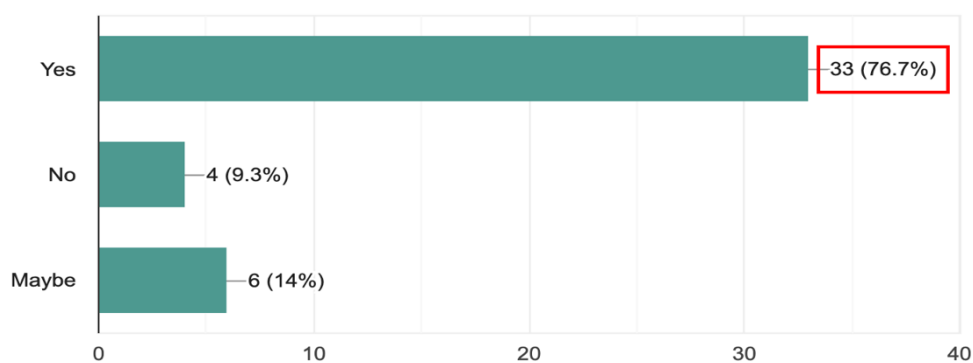

Some students reported that the graphs helped them set goals, either to participate more actively and increase their turn-taking (47.6%), or to stop monopolising the floor time and allow others to speak (7.1%) (see Figure 3). An important finding of our study is that students (64.3%) found that the L2L visualisations mainly enhanced their confidence in their L2 interaction with native speakers. According to the students, in previous teletandem projects, they often came out of the interactions thinking that they had not participated much in French, but thanks to L2L, they were pleasantly surprised to discover that their interaction was quite high in French too.

Figure 3: Learner feedback on how L2L proved helpful

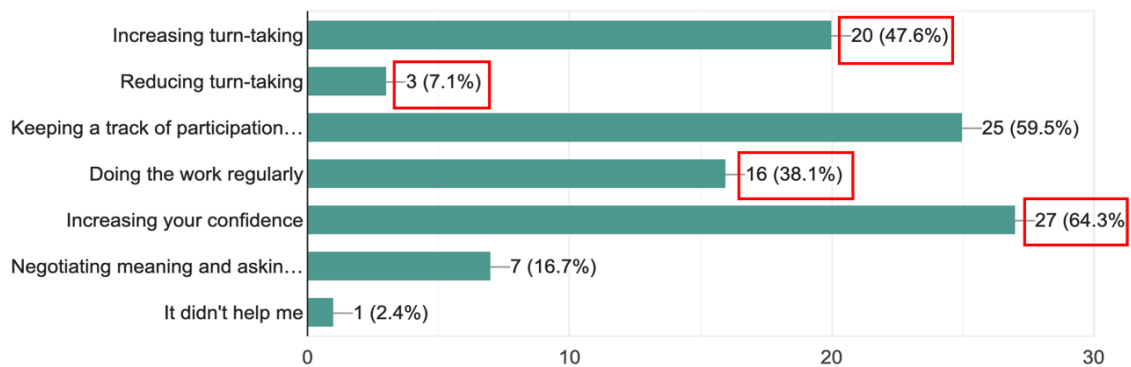

The four most recurring benefits of the visualisations reported in students' qualitative feedback were getting weekly visual feedback on how much they participated; keeping a check on their engagement and setting objectives to improve in following sessions; finding the motivation to improve each week; and gaining confidence and regulating their interaction in following instantiations. About 38% of respondents (Figure 3) felt that L2L helped them to come out of their comfort zones and engage in their synchronous interactions regularly instead of avoiding them as was the case in previous teletandem projects, as students' full interactions could now be easily monitored by the lecturer.

With regard to whether the L2L system helped students improve their language learning experience or not, a student reflected:

> "Definitely! At first I didn't understand the point of it but as I'm currently rewatching our sessions, I can easily identify the parts where I spoke the most/least and identify the reason why".

L2L, therefore, acted as a means of empowering language learners by inciting them "to make conscious decisions about what they can do to improve their learning" (Anderson, 2008, p. 99). This in turn helped them to work on the quality of their interactions with native speakers. In general, students felt that L2L made the interactions more interactive, and although students mentioned that it helped them notice grammar mistakes, most students thought that it helped with interaction skills, such as long pauses, hesitations, poor backchanneling, low or high participation, and metacognitive awareness to understand why this was happening and how it could be overcome, rather than language learning per se.

## 4. Conclusions

L2L was deemed useful by students to get weekly visual feedback, keep track of their participation, and set objectives for the next session, but mainly to gain confidence and regulate the quantity of their interactions in following instantiations. The automated quantitative and qualitative data allowed students to engage in reflection as part of the pedagogical design and set objectives for themselves. The video playback function allowed students to focus on the quality of the synchronous content although this was not exploited fully. A possible future step could be the organisation of co-reflections between students using the playback function to facilitate peer-review.